\documentclass[10pt,twocolumn,letterpaper]{article}

\usepackage{iccv}
\usepackage{times}
\usepackage{epsfig}
\usepackage{graphicx}
\usepackage{amsmath}
\usepackage{amssymb}

\usepackage{bbm}
\usepackage{algorithm}
\usepackage{algorithmic}
\usepackage{subcaption}
\usepackage{graphicx}
\usepackage{amsmath}
\usepackage{amssymb}
\usepackage{booktabs}
\usepackage{multirow}
\usepackage{nicefrac}

\DeclareMathOperator*{\onehot}{onehot}

\DeclareMathOperator*{\argmax}{arg\,max}

\usepackage{bm}

\usepackage{xcolor}
\definecolor{myblue}{RGB}{46, 117, 182}
\definecolor{myred}{RGB}{237, 125, 49}
\definecolor{myyellow}{RGB}{255, 192, 0}
\definecolor{mygreen}{RGB}{0, 176, 80}

\usepackage[pagebackref=true,breaklinks=true,letterpaper=true,colorlinks,bookmarks=false]{hyperref}

\iccvfinalcopy 

\def\iccvPaperID{2637} 

\ificcvfinal\fi

\begin{document}

\title{Learning to Select Camera Views: \\ Efficient Multiview Understanding at Few Glances}

\author{Yunzhong Hou \qquad Stephen Gould \qquad Liang Zheng \\
Australian National University\\
{\tt\small \{firstname.lastname\}@anu.edu.au}
}

\maketitle
\ificcvfinal\thispagestyle{empty}\fi

\begin{abstract}
Multiview camera setups have proven useful in many computer vision applications for reducing ambiguities, mitigating occlusions, and increasing field-of-view coverage. However, the high computational cost associated with multiple views poses a significant challenge for end devices with limited computational resources. To address this issue, we propose a view selection approach that analyzes the target object or scenario from given views and selects the next best view for processing. Our approach features a reinforcement learning based camera selection module, MVSelect, that not only selects views but also facilitates joint training with the task network. Experimental results on multiview classification and detection tasks show that our approach achieves promising performance while using only 2 or 3 out of $N$ available views, significantly reducing computational costs. Furthermore, analysis on the selected views reveals that certain cameras can be shut off with minimal performance impact, shedding light on future camera layout optimization for multiview systems. Code is available at \url{https://github.com/hou-yz/MVSelect}.

\end{abstract}

\section{Introduction}
\label{sec:intro}

\begin{figure}
\centering
\includegraphics[width=0.95\linewidth]{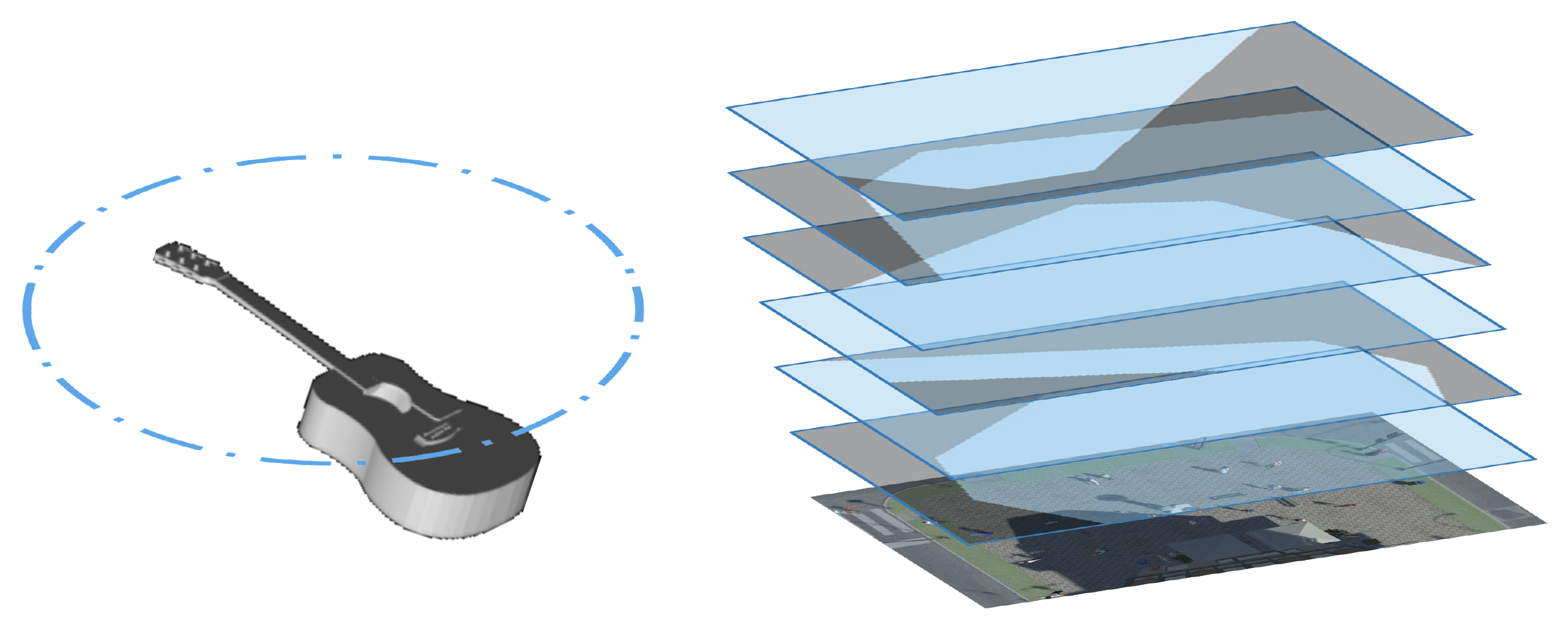}
\caption{
Example of multiview camera setups. \textbf{Left}: multiview classification jointly considers multiple camera views to identify the object. Cameras (\textcolor{myblue}{blue dots}) follow a certain layout \cite{wu20153d,kanezaki2018rotationnet} to provide complementary viewpoints. \textbf{Right}: multiview detection estimates pedestrian occupancy from bird's-eye-view (bottom colored image). Cameras (\textcolor{myblue}{blue FoV maps}) can be set at different locations \cite{chavdarova2018wildtrack,hou2020multiview} to increase FoV coverage and deal with occlusion. 
For both classification and detection tasks, due to hardware constraints, camera layouts are usually pre-defined. 
}
\label{fig:intro}
\end{figure}


Multiple camera views (multiview) are popular in computer vision systems for their ability to address challenges such as occlusions, ambiguities, and limited field-of-view (FoV) coverage. Tasks like classification \cite{su2015multi,qi2016volumetric} and detection \cite{chavdarova2018wildtrack,hou2020multiview} have shown significant benefits from using multiple cameras (Fig.~\ref{fig:intro}). With reduced hardware cost and easy deployment, real-world products now include more cameras at larger scales. For example, iPhones went from one to three rear lenses \cite{iphone14pro}; autopilot cars went from two cameras in the ARGO project \cite{argo} to eight in Tesla \cite{tesla_model3_manual}; and future smart cities will also incorporate more cameras \cite{smart_city_camera}.

However, the use of multiple cameras comes at a high computational cost, which can be a significant challenge for end devices with limited computational resources, especially with higher image resolutions \cite{shermeyer2019effects,sabottke2020effect} and deeper neural network backbones \cite{he2016deep,dosovitskiy2020vit}. Limiting image resolution or using lighter networks \cite{molchanov2016pruning,howard2017mobilenets} are current options to reduce computation, but they may impede the progress in camera sensors or neural network architecture. 

\begin{figure}
\centering
\includegraphics[width=0.85\linewidth]{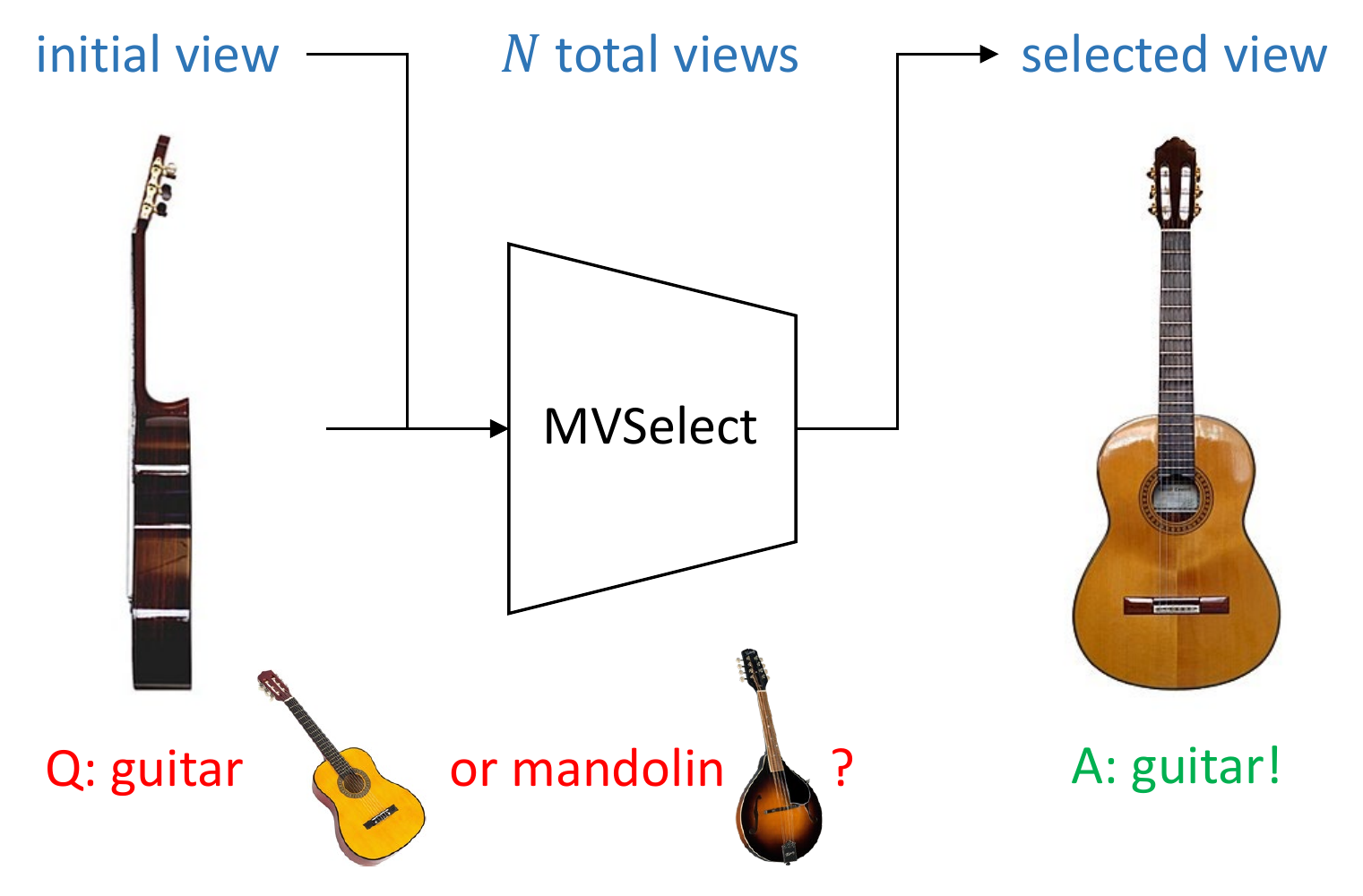}
\caption{
Efficient multiview understanding with two glances. Instead of using all $N$ views at once, a more efficient approach is to first examine one view and then select another view that should resolve ambiguities in the initial glance. In this example, if the initial side view cannot distinguish between a guitar and a mandolin (a round-shaped instrument that may also have a flat back), we can query the front view in the second glance to clarify the ambiguities. 
}
\label{fig:demo}
\end{figure}

To address this challenge, this paper proposes a new approach to efficient multiview understanding by selecting only the most useful views. To identify the best views, this approach leverages  camera layouts, which is a key aspect overlooked by existing alternatives. With known camera layouts, networks should be able to infer what each camera view looks like and then choose accordingly, as previous works \cite{kanezaki2018rotationnet} have shown that networks can associate images with camera poses. Existing work on active vision  \cite{aloimonos1988active,findlay2003active,chen2011active} also indicates that it is possible to smartly select the camera locations and directions for better robot vision. In our problem, for example (see Fig.~\ref{fig:demo}), we might confuse a guitar with a mandolin from the side view. Previous knowledge of the camera layout should inform the system that the front view can clear the ambiguities and should be queried. 



To achieve this goal, this paper proposes a novel view selection module, MVSelect, that chooses the best camera views from any initial view. The proposed module first analyzes the target object or scenario using the given view and then selects the next view that best helps the task (multiview classification or detection) network. To navigate through the non-differentiable parts in view selection, we formulate this interactive process as a reinforcement learning problem \cite{mnih2014recurrent} that considers the chosen cameras and their visual feedback as state, the next camera to select as action, and the final result from the task network as reward. 
%

On both multiview classification and detection tasks, our experimental results show that MVSelect can provide a good strategy for fixed task networks, while also capable of joint training with the task network for further performance improvements. Specifically, when allowing joint training on both MVSelect and the task network, the resulting system can achieve similar results to using all $N$ cameras, while using only 2 views for classification tasks and 3 views for detection tasks, respectively.


The computational overhead of MVSelect is very small, as it shares the feature extraction backbone with the task network and only has a few learnable layers. For the entire system, the computational cost is roughly proportional to the number of views used, resulting in approximately $\nicefrac{2}{N}$ of the total computation when only 2 views are used, a significant efficiency boost. 

The MVSelect policy also enables study on multiview camera layout. In fact, we find that many of the $N$ cameras are rarely chosen and can be shut off for further operational cost improvements, which can serve as a starting point for future study on multiview camera layout.




\section{Background}
\label{sec:related work}

\textbf{Multiview classification.} One effective way for 3D shape recognition is to capture the object in multiple camera views. MVCNN \cite{su2015multi} extracts feature vectors from the input views, and then uses max pooling to aggregate across multiple views for classification. Based on MVCNN, many alternative approaches are proposed. Qi \etal \cite{qi2016volumetric} propose sphere rendering at different volume resolutions. GVCNN \cite{feng2018gvcnn} investigates hierarchical information between different views by grouping the image features before the final aggregation. RotationNet \cite{kanezaki2018rotationnet} introduces a multi-task objective by jointly considering classification and camera poses. ViewGCN \cite{wei2020view} uses a Graph Convolution Network (GCN) \cite{kipf2016semi} instead of the max pooling layer to aggregate across views. Recently, Hamdi \etal \cite{hamdi2021mvtn} propose the MVTN network to estimate the best viewpoints for 3D point cloud models.

\textbf{Multiview detection.} Occlusion is a key problem for object detection, and is by nature very difficult to tackle using only one camera view. To deal with this problem, researchers investigate multiview approaches for pedestrian detection and estimate occupancy from the bird's-eye-view (BEV). For this task, some methods \cite{fleuret2007multicamera,roig2011conditional,xu2016multi} aggregate single-view detection results. Others \cite{baque2017deep,chavdarova2017deep,hou2020multiview} find single-view detection results unreliable and instead aggregate the features. Specifically, Hou \etal \cite{hou2020multiview} introduce MVDet, a fully convolutional approach that projects feature maps from each camera view to the BEV. Inspired by  MVDet, researchers introduce other novel methods. SHOT \cite{song2021stacked} projects the image feature map at different heights and stacks them together to improve performance. MVDeTr \cite{hou2021multiview} deals with distinct distortion patterns from the projection. 
Qiu \etal \cite{qiu20223d} investigate data augmentation with simulated occlusion over multiple views.

\textbf{Camera viewpoint study.} Camera viewpoint is an important factor in multiview systems. Previous works on active vision investigate sensor planning, where the pose and settings of vision sensors are determined for robot vision tasks \cite{aloimonos1988active,findlay2003active,chen2011active}. In multiview classification, RotationNet \cite{kanezaki2018rotationnet} makes some pioneering work on limited view numbers by taking a random partial set of all $N$ views of the image. MVTN \cite{hamdi2021mvtn} uses the 3D point cloud as initial input and then estimates the best camera layout for multiview classification, but its moving camera assumption is hard to met in real-world systems such as iPhones \cite{iphone14pro} and Teslas \cite{tesla_model3_manual}. In multiview detection, Vora \etal \cite{vora2021bringing} investigate camera layout generalization by randomly dropping camera views from both training and testing. 

\textbf{Reinforcement learning} (RL) directs an agent to interact with the environment in a manner that maximizes cumulative rewards. State $s \in \mathcal{S}$, action $a \in \mathcal{A}$, and reward $r \in \mathcal{R}$ are key concepts to model the interaction between agent and environment. In a certain state $s$, policy $\pi\left(a|s\right)$ records the probability for each action, and state value function $V\left(s\right)$ estimates the future rewards when following the corresponding policy. To learn the best policy, Q-learning and DQN \cite{mnih2013playing} optimize the action value function $Q\left(s, a\right)$, which describes the estimated future return for a specific action $a$ at state $s$. Policy gradient methods like REINFORCE \cite{williams1992simple} and PPO \cite{schulman2017proximal} directly optimize for the polity $\pi\left(a|s\right)$. 

Reinforcement learning has been used in  many vision tasks. To deal with non-differentiable parts, Schulman \etal \cite{schulman2015gradient} use RL to estimate loss function's gradient. Minh \etal \cite{mnih2014recurrent} recognize an image by navigating an agent with a small receptive field. Embodied vision \cite{zhu2017target, anderson2018vision} uses RL to navigate an agent to the final destination. 

\begin{figure}
\centering
\includegraphics[width=\linewidth]{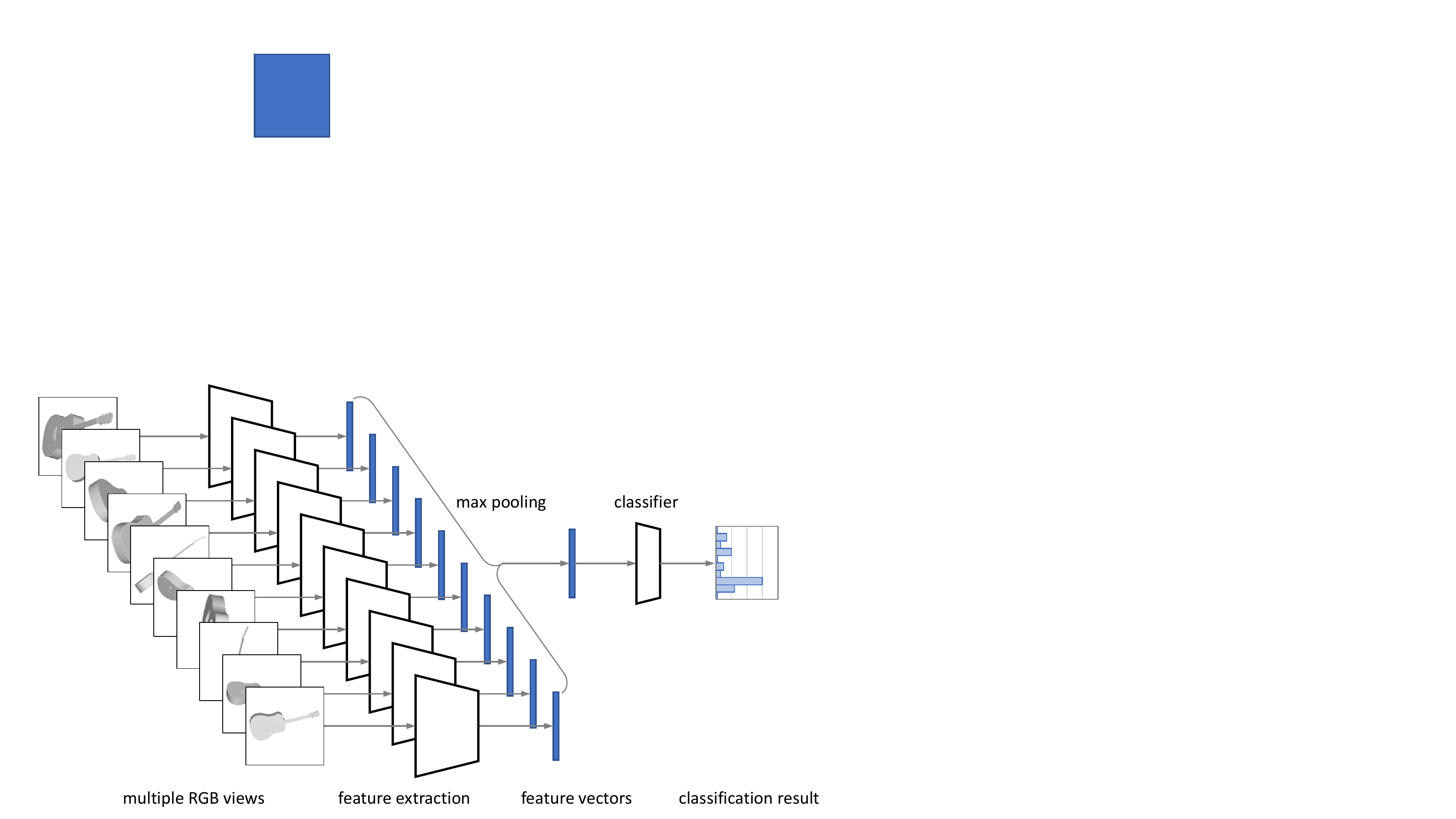}
\caption{
Multiview classification with MVCNN \cite{su2015multi}. 
}
\vspace{-3mm}
\label{fig:mvcnn}
\end{figure}


\section{Multiview Network Revisit}
\label{sec:revisit}

\subsection{Multiview Classification: MVCNN}
\label{secsec:mvcnn}

MVCNN \cite{su2015multi} (Fig.~\ref{fig:mvcnn}) is a classic architecture which many multiview classification networks \cite{qi2016volumetric,feng2018gvcnn,kanezaki2018rotationnet,yu2018multi,wang2019dominant,yang2019learning,wei2020view,hamdi2021mvtn} build upon. Given $N$ input images $\bm{x}_n, n\in\left\{1,\ldots,N\right\}$, first, MVCNN uses its feature extractor $f\left(\cdot\right)$ to calculate the feature vectors, 
\begin{align}
\label{eq:feature extraction}
    \bm{h}_{n} = f\left(\bm{x}_n\right),
\end{align}
where the feature vector $\bm{h}_n\in\mathbb{R}^{D}$ is $D$-dimensional. Secondly, it uses max pooling to aggregate multiple views into an overall feature descriptor $\hat{\bm{h}}\in\mathbb{R}^{D}$,
\begin{align}
\label{eq:multiview aggregation}
    \hat{\bm{h}}_d = \max_{n}{\left\{\bm{h}_{n,d}\right\}}, \quad d\in\left\{1,\ldots,D\right\},
\end{align}
where $\hat{\bm{h}}_d$ and $\bm{h}_{n,d}$ denote the $d$ dimension of $\hat{\bm{h}}$ and $\bm{h}_{n}$, respectively. 
Lastly, it applies the output head $g\left(\cdot\right)$ as,
\begin{align}
\label{eq:output head}
    \hat{\bm{y}} = g\left(\hat{\bm{h}}\right),
\end{align}
to produce the classification result $\hat{\bm{y}}$. 


In training, the original design by Su \etal \cite{su2015multi} adopts a 2-stage paradigm by first training on individual views and then considering multiple views. In this paper, we skip the first stage and directly train MVCNN on all $N$ views,
\begin{align}
\label{eq:loss mvcnn}
    \mathcal{L}_\text{MVCNN} = \mathcal{L}_\text{CE}\left(\hat{\bm{y}}, \bm{y}\right),
\end{align}
where $\mathcal{L}_\text{CE}\left(\cdot,\cdot\right)$ denotes the cross-entropy loss and $\bm{y}$ denotes the ground truth one-hot label.

\begin{figure}
\centering
\includegraphics[width=\linewidth]{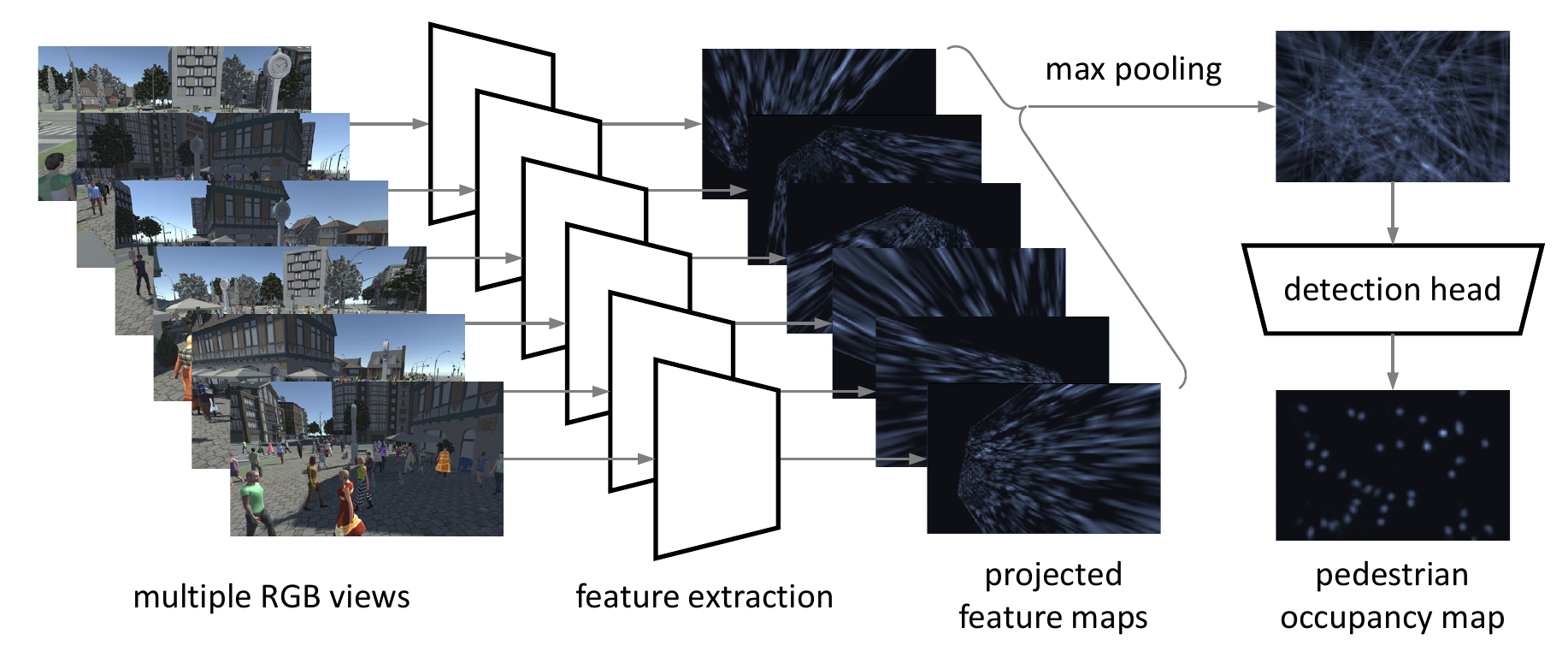}
\caption{
Multiview detection with MVDet \cite{hou2020multiview}. 
}
\vspace{-3mm}
\label{fig:mvdet}
\end{figure}


\subsection{Multiview Detection: MVDet}
\label{secsec:mvdet}

MVDet \cite{hou2020multiview} (Fig.~\ref{fig:mvdet}) is a multiview detection architecture that estimates human occupancy in bird's-eye-view (BEV). Its feature projection and anchor-free design have been adopted by many recent works \cite{hou2021multiview,song2021stacked,vora2021bringing,qiu20223d,gao2022exploiting,lima20223d,hwang2022booster}. Given input images $\bm{x}_n, n\in\left\{1,\ldots,N\right\}$, MVDet first extracts $D$-channel feature maps for each view and uses perspective transformation to project the camera views to the BEV. These operations can be jointly considered as the BEV feature extraction step under Eq.~\ref{eq:feature extraction}, with the exception that $\bm{h}_n\in\mathbb{R}^{D\times H \times W}$ now denotes the $D$-channel feature map for the BEV scenario of shape $H\times W$. Secondly, for multiview aggregation, instead of the concatenation in the original design, we choose element-wise max pooling in Eq.~\ref{eq:multiview aggregation}, producing the overall feature description $\hat{\bm{h}}\in\mathbb{R}^{D\times H \times W}$ that fits arbitrary numbers of views. Lastly, we apply the output head as Eq.~\ref{eq:output head} to generate a heatmap $\hat{\bm{y}}\in\left[0,1\right]^{H\times W}$ that indicates the likelihood of human occupancy in each BEV location.




The loss we used for MVDet training can be written as,
\begin{align}
\label{eq:loss mvdet}
    \mathcal{L}_\text{MVDet} = \mathcal{L}_\text{BEV}\left(\hat{\bm{y}}, \bm{y}\right) + \frac{1}{N}\sum_{n=1}^{N}{\mathcal{L}_n},
\end{align}
where $\mathcal{L}_\text{BEV}\left(\cdot,\cdot\right)$ denotes BEV output loss; $\bm{y}\in \left\{0,1\right\}^{H\times W}$ denotes binary ground truth map; and $\mathcal{L}_n$ denotes the auxiliary per-view loss with 2D bounding boxes.

\begin{figure*}
\centering
\includegraphics[width=\linewidth]{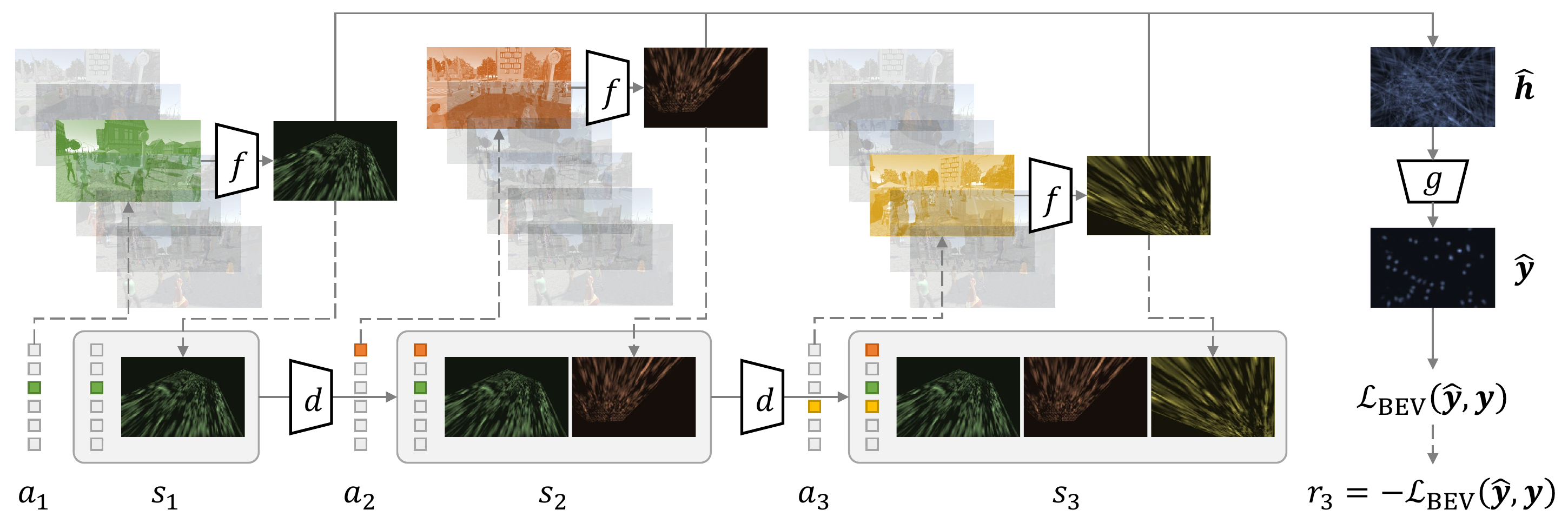}
\caption{
Efficient multiview understanding with a total of $T=3$ glances. Solid lines indicate the network forward pass and dashed lines indicate the interaction between \textit{agent} (MVSelect) and \textit{environment} (multiview system). The approach starts with a random \textcolor{mygreen}{initial view} $a_1$, and the feature extractor $f\left(\cdot\right)$ is used to compute its feature $\bm{h}_{a_1}$ (Eq.~\ref{eq:feature extraction}). The \textit{state} for this initial time step is recorded as $s_1$ (Eq.~\ref{eq:state}). Next, the proposed MVSelect module $d\left(\cdot\right)$ is used to choose a \textcolor{myred}{second view} $a_2$, which is the \textit{action} for \textit{state} $s_1$. The feature representation $\bm{h}_{a_2}$ for this second view is computed, and the state for the two views is updated as $s_2$. Then, by repeating the last step, a \textcolor{myyellow}{third view} $a_3$ is chosen based on state $s_2$. The feature representation $\bm{h}_{a_3}$ and the state $s_3$ are also updated accordingly. Finally, three views are aggregated into an \textcolor{myblue}{overall} descriptor $\hat{\bm{h}}$ (Eq.~\ref{eq:multiview aggregation}), and the task network output $\hat{\bm{y}}$ is calculated using the output head $g\left(\cdot\right)$ (Eq.~\ref{eq:output head}). For multiview detection, the final \textit{reward}  $r_3=-\mathcal{L}_\text{BEV}\left(\hat{\bm{y}}, \bm{y}\right)$  is set as the negative of the BEV loss (Eq.~\ref{eq:reward}), and all other rewards are assigned as zero. 
}
\label{fig:method}
\end{figure*}

\section{Efficient Multiview Understanding}
\label{sec:method}

In a multiview system with $N$ views, our efficient approach uses a total of $T<N$ camera views $a_t\in\left\{1,\ldots,N\right\}, t\in\left\{1,\ldots,T\right\}$ for multiview understanding. To achieve this, we propose a view selection module, MVSelect, denoted as $d\left(\cdot\right)$, which sequentially selects camera views. Starting from a random initial view $a_1$, MVSelect chooses the remaining $T-1$ cameras by observing the target object or scene from existing views $a_1, \ldots, a_t$ at each time step $t$ and deciding which camera view $a_{t+1}$ to select next. The resulting $T$ views should give the task network high classification or detection performance. 

Once $T$ camera views have been gathered, we aggregate them into an overall description $\hat{\bm{h}}$ using Eq.~\ref{eq:multiview aggregation}, and calculate the final output $\hat{\bm{y}}$ using Eq.~\ref{eq:output head}. Figure~\ref{fig:method} gives an overview of the proposed efficient multiview approach.



\subsection{Problem Formulation}
\label{secsec:formualtion}
We can frame this interactive process as a reinforcement learning problem, where MVSelect is the \textit{agent} and the multiview system is the \textit{environment}.

\textbf{State.} For a time step $t$, we record the chosen camera views $a_1, \ldots, a_t$ and the observations $\bm{h}_{a_1}, \ldots, \bm{h}_{a_t}$ as state $s_t$. We use the extracted features to represent the observations rather than the RGB camera views $\bm{x}_{a_1}, \ldots, \bm{x}_{a_t}$ to reduce dimensionality and maximize efficiency, since these features will be used in the task network later (Eq.~\ref{eq:multiview aggregation}).

Mathematically, we formulate the state $s_t$ as follows,
\begin{equation}
\begin{gathered}
\label{eq:state}
s_t = \left<s^\text{cam}_t, s^\text{obs}_t\right>, \\
s^\text{cam}_t = \sum_{\tau=1}^{t}{\onehot\left(a_{\tau}\right)}, \\
s^\text{obs}_{t,d} = \max_{\tau=1}^{t}{\left\{\bm{h}_{a_{\tau},d}\right\}}, \; d\in\left\{1,\ldots,D\right\},
\end{gathered}
\end{equation}
where $\onehot\left(\cdot\right)$ is the one-hot function over $N$ cameras. This representation reflects both chosen cameras $s^\text{cam}_t \in \mathbb{R}^N$ and their observations $s^\text{obs}_t \in \mathbb{R}^D$, and maintains the same dimensionality across different time steps. The observation part $s^\text{obs}_t$ also matches the overall representation in Eq.~\ref{eq:multiview aggregation}.


\textbf{Action.} 
For state $s_t, t\in\left\{1,\ldots,T-1\right\}$, MVSelect takes the $t+1$ camera view $a_{t+1}$ as action. 

\textbf{Reward.} 
Upon taking action $a_{t+1}$, the system receives reward $r_{t+1}$ and transitions into the next state $s_{t+1}$. 
To achieve high task network performance, we consider the following as reward,
\begin{equation}
\begin{gathered}
\label{eq:reward}
r_t=0, t\in\left\{1,\ldots,T-1\right\}, \\
r^\text{MVCNN}_T = \mathbbm{1}\left(\hat{\bm{y}}=\bm{y}\right),  \;
r^\text{MVDet}_T = -\mathcal{L}_\text{BEV}\left(\hat{\bm{y}},\bm{y}\right),\\
\end{gathered}
\end{equation}
where $\mathbbm{1}\left(\cdot\right)$ denotes the binary indicator function. For the last time step, we use classification accuracy or the inverse of the BEV loss (Eq.~\ref{eq:loss mvdet}) as reward $r_T$ for the two problems, respectively. For all other time steps, we set $r_t$ as zero.



\begin{figure}
\centering
\includegraphics[width=0.9\linewidth]{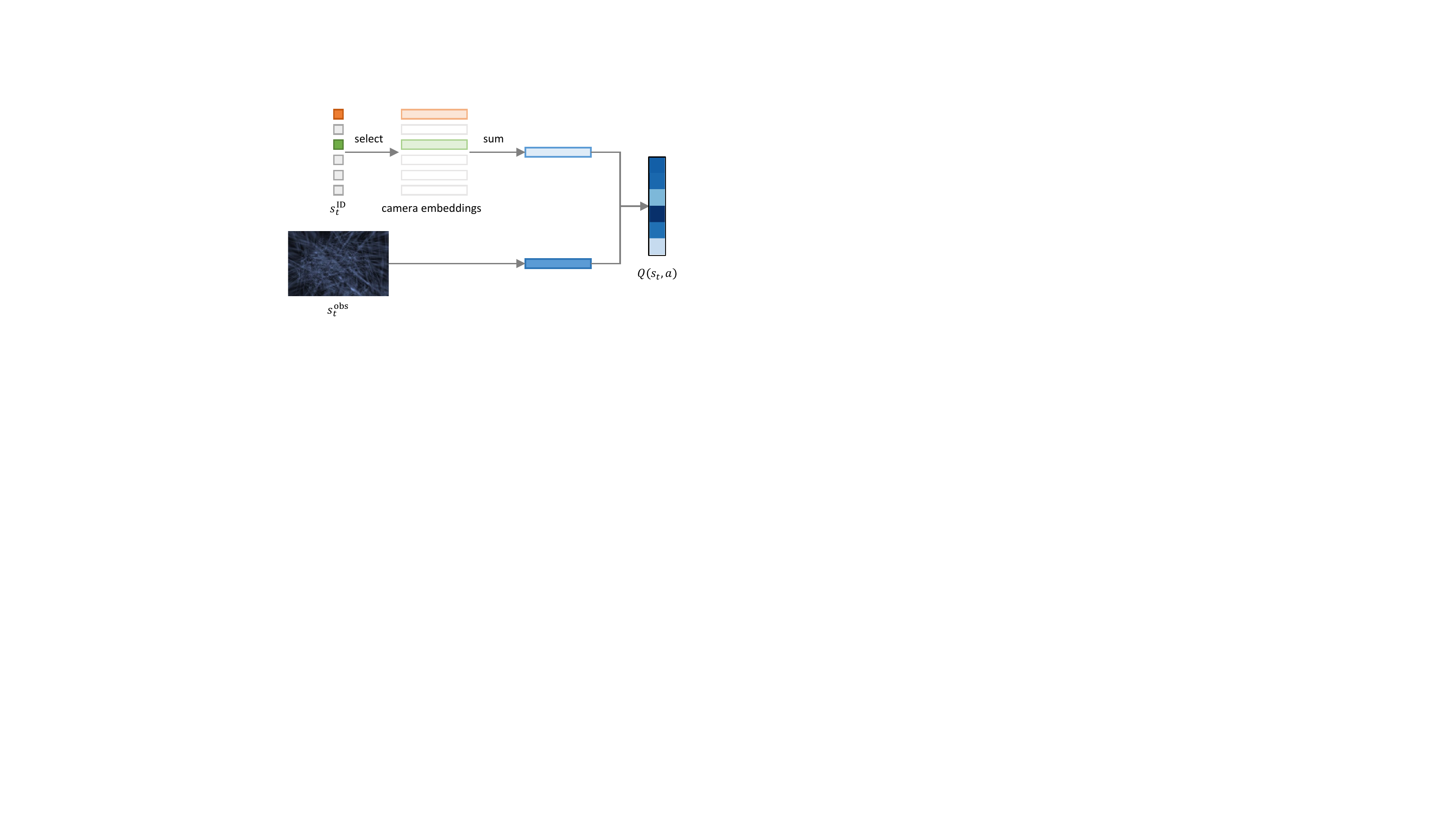}
\caption{
MVSelect architecture. 
}
\label{fig:mvselect}
\end{figure}

\subsection{MVSelect Architecture}
\label{secsec:mvselect}

As shown in Fig.~\ref{fig:mvselect}, we design MVSelect architecture $d\left(\cdot\right)$ with two branches. The first branch expands the camera selection result $s_t^\text{cam} \in \mathbb{R}^N$ into $D$-dimensional learnable camera embeddings, and then sums over the selected embeddings to formulate a hidden vector. The second branch processes  the observation $s_t^\text{obs} \in \mathbb{R}^D$, and converts that into another  hidden vector. By combining the two hidden vectors, MVSelect outputs the action-value $Q\left(s, a\right)$, which measures the expected cumulative rewards for taking an action $a$ in a given state $s$. 

During testing, MVSelect outputs the next action as,
\begin{equation*}
\label{eq:mvselect}
    a_{t+1} = \argmax_{a}{Q\left(s_t, a\right)},
\end{equation*}
which maximizes the expected cumulative rewards.





\subsection{Training Scheme}
\label{secsec:training}

We adopt Q-learning  \cite{sutton2018reinforcement} for training MVSelect. Specifically, action-value function $Q\left(\cdot,\cdot\right)$ should estimate the cumulative future rewards after taking action $a_{t+1}$ at state $s_t$,
\begin{equation*}
    Q\left(s_t, a_{t+1}\right)= \mathbb{E}\left(\sum_{\tau=t+1}^{T}{\gamma^{\tau-t-1}r_{\tau}}\right),
\end{equation*}
where $\mathbb{E}\left(\cdot\right)$ denotes the expectation, and $\gamma\in \left[0,1\right]$ denotes the discount factor. We take the temporal difference (TD) \cite{sutton2018reinforcement} target as supervision for the action value, 
\begin{equation}
\label{eq:td target}
    q_t=\begin{cases}
    r_{t+1} + \gamma \max_{a}{Q\left(s_{t+1}, a\right)},& \text{if } t < T-1\\
    r_{T},              & \text{otherwise}
    \end{cases},
\end{equation} 
and calculate the loss using the $L_2$ distance,
\begin{equation}
\label{eq:loss q}
    \mathcal{L}_\text{RL} = \sum_{t=1}^{T-1}{\mathcal{L}_\text{MSE}\left(Q\left(s_t, a_{t+1}\right), q_t\right)},
\end{equation}
where the next action $a_{t+1}$ is chosen using $\epsilon$-greedy for exploration-exploitation trade offs (see Algorithm~\ref{alg:joint training}). 

In joint training, the task network takes supervision from the task loss $\mathcal{L}_\text{task}$ (Eq.~\ref{eq:loss mvcnn} and Eq.~\ref{eq:loss mvdet}), and the selection module takes supervision from the RL loss $\mathcal{L}_\text{RL}$ (Eq.~\ref{eq:loss q}). We demonstrate this process step-by-step in Algorithm~\ref{alg:joint training}.

\begin{algorithm}[t]
\caption{Joint training of MVSelect and task network.}
\label{alg:joint training}
\begin{algorithmic}[1]
\small
\STATE \textbf{input}: camera views $\bm{x}_n, n \in \left\{1,\ldots,N\right\}$, ground truth label $\bm{y}$, random initial view $a_1$, number of total views $T$, $\epsilon$, $\gamma$. 
\STATE initialize the total loss $\mathcal{L}_\text{total}=0$;
\FOR{$t \in \left\{1,\ldots,T-1\right\}$}
    \STATE select and apply the next action using $\epsilon$-greedy: with probability $\epsilon$ adopt a random action, or else choose the action with highest value $a_{t+1} = \argmax_{a}{Q\left(s_t, a\right)}$;
    \STATE observe next state $s_{t+1}$ (Eq.~\ref{eq:state}) and reward $r_{t+1}$ (Eq.~\ref{eq:reward});
    \STATE calculate the TD \cite{sutton2018reinforcement} target $q_t$ (Eq.~\ref{eq:td target}) and update RL loss $\mathcal{L}_\text{total}=\mathcal{L}_\text{total}+\mathcal{L}_\text{RL}$ (Eq.~\ref{eq:loss q});
\ENDFOR
\STATE calculate the task loss $\mathcal{L}_\text{task}$ (Eq.~\ref{eq:loss mvcnn} and Eq.~\ref{eq:loss mvdet});
\STATE optimize for the total loss $\mathcal{L}_\text{total}=\mathcal{L}_\text{total}+\mathcal{L}_\text{task}$;
\STATE \textbf{update}: task network  $f\left(\cdot\right)$ and $g\left(\cdot\right)$, and MVSelect $d\left(\cdot\right)$.
\end{algorithmic}
\end{algorithm}

\section{Experiments}
\label{sec:experiments}

\subsection{Experiment Settings}
\textbf{Datasets.} 
We verify the performance of the proposed approach on multiview classification and detection tasks. 


\textit{ModelNet40} is a subset of  3D CAD models in ModelNet \cite{wu20153d}. It includes 40 categories of synthetic 3D objects with 9,843 training models and 2,468 test models. For multiview classification experiments, we use two different configurations: the \textit{12-view} circular configuration from MVCNN \cite{su2015multi} and the \textit{20-view} dodecahedral configuration from RotationNet \cite{kanezaki2018rotationnet}.

\begin{table*}[]
\footnotesize
\begin{tabular}{l|c|c}
\toprule
             & \multicolumn{2}{c}{ModelNet40 \cite{wu20153d}} \\ \cline{2-3}
             & 12 views & 20 views \\ \hline
MVCNN \cite{su2015multi}        & 90.1     & 92.0       \\ 
GVCNN \cite{feng2018gvcnn}       & 92.6     & -        \\ 
MHBN \cite{yu2018multi}     & 93.4     & -        \\ 
RotationNet \cite{kanezaki2018rotationnet} & -        & 94.7     \\ 
RelationNet \cite{yang2019learning}     & 94.3        & 97.3     \\ 
ViewGCN \cite{wei2020view}     & -        & \textbf{97.6}     \\ 
MVTN* \cite{hamdi2021mvtn}       & 93.8     & 93.5     \\ \hline
MVCNN (ours) & \textbf{94.5}     & 96.5     \\ \bottomrule
\end{tabular}
\hfill
\footnotesize
\begin{tabular}{l|cccc|cccc}
\toprule
               & \multicolumn{4}{c|}{Wildtrack \cite{chavdarova2018wildtrack}} & \multicolumn{4}{c}{MultiviewX \cite{hou2020multiview}} \\ \cline{2-9}
               & MODA  & MODP  & prec.    & recall    & MODA   & MODP  & prec.  & recall  \\ \hline
RCNN \&   cluster \cite{xu2016multi} & 11.3  & 18.4  & 68    & 43    & 18.7   & 46.4  & 63.5  & 43.9  \\
POM-CNN \cite{fleuret2007multicamera}             & 23.2  & 30.5  & 75    & 55    & -      & -     & -     & -     \\
DeepMCD \cite{chavdarova2017deep}       & 67.8  & 64.2  & 85    & 82    & 70     & 73    & 85.7  & 83.3  \\
Deep-Occlusion \cite{baque2017deep} & 74.1  & 53.8  & 95    & 80    & 75.2   & 54.7  & 97.8  & 80.2  \\
MVDet \cite{hou2020multiview}         & 88.2  & 75.7  & 94.7  & 93.6  & 83.9   & 79.6  & 96.8  & 86.7  \\
SHOT \cite{song2021stacked}          & 90.2  & 76.5  & 96.1  & 94.0    & 88.3   & 82.0    & 96.6  & 91.5  \\
MVDeTr \cite{hou2021multiview}        & \textbf{91.5}  & \textbf{82.1}  & \textbf{97.4}  & 94.0    & \textbf{93.7}   & \textbf{91.3}  & \textbf{99.5}  & 94.2  \\ \hline
MVDet (ours)   & 90.0  & 80.9  & 95.4  & \textbf{94.5}  & 93.0   & 90.3  & 98.7  & \textbf{94.4} \\ \bottomrule
\end{tabular}
\caption{Performance comparison with state-of-the-art multiview classification and multiview detection methods. Results are averaged from 5 runs. * indicates that the camera poses are dynamically chosen and do not follow a pre-defined layout. }
\label{tab:sota}
\end{table*}



\textit{Wildtrack} \cite{chavdarova2018wildtrack} is a real-world multiview detection dataset with 7 camera views covering a $12 \times 36$ square meter area, which is represented as a $480 \times 1440$ grid from BEV. It contains 360 frames for training and 40 frames for testing. 

\textit{MultiviewX} \cite{hou2020multiview} is a synthetic multiview detection dataset created using the Unity \cite{unity} engine. It has 6 cameras with higher pedestrian density than Wildtrack. It focuses on a $16 \times 25$ square meter area, which is discretized into $640 \times 1000$ BEV grid. Like Wildtrack, MultiviewX also contains 360 training frames and 40 testing frames.

\textbf{Evaluation metrics.}
For multiview classification, we follow previous methods \cite{qi2016volumetric,kanezaki2018rotationnet,wei2020view,yu2018multi,yang2019learning,hamdi2021mvtn} and report instance-averaged accuracy as the primary indicator. 

Regarding multiview detection, we report the following metrics: multi-object detection accuracy (MODA), multi-object detection precision (MODP), precision, and recall \cite{kasturi2008framework}. During evaluation, we first compute false positives (FP), false negatives (FN), and true positives (TP), and then use them to calculate the metrics. Specifically, MODA is calculated as $1-\frac{\text{FP}+\text{FN}}{\text{GT}}$, where GT is the number of ground truth pedestrians. MODP is calculated as $\frac{\sum{1-\text{dist}[\text{dist}<\text{thres}]/\text{thres}}}{\text{TP}}$, where $\text{dist}$ is the distance from the estimated pedestrian location to its ground truth and $\text{thres}$ is the threshold of 0.5 meters. MODP indicates the BEV localization accuracy. Precision and recall are calculated as $\frac{\text{TP}}{\text{TP}+\text{FP}}$ and $\frac{\text{TP}}{\text{GT}}$, respectively. 


All metrics are reported in percentages.

\textbf{Implementation details.}
For multiview classification, we input images of size $224\times 224$ to the MVCNN model. For multiview detection, we use a resolution of $720 \times 1280$ for input images with view-coherent data augmentation \cite{hou2021multiview}, and downsample the BEV grid by a factor of $4$. In terms of architecture, we use ResNet-18 \cite{he2016deep} as feature extractor $f\left(\cdot\right)$ for both tasks. 

We train all networks for 10 epochs using the Adam optimizer \cite{kingma2015adam}. We use learning rates of $5\times10^{-5}$ and $5\times10^{-4}$, with batch sizes of 8 and 1 for MVCNN and MVDet, respectively. The MVSelect module is trained using a learning rate of $1\times10^{-4}$. For joint training, we decrease the learning rate for the task network to $\nicefrac{1}{5}$ of its original value. 

Regarding hyperparameters, we set the future reward discount factor $\gamma=0.99$, and the exploration ratio $\epsilon$ to gradually decrease from $0.95$ to $0.05$ during training. 

All experiments are conducted on a single RTX-3090 GPU and averaged across 5 repetitive runs.

\subsection{Evaluation of Task Networks}
In Table \ref{tab:sota}, we compare our implementations of MVCNN \cite{su2015multi} and MVDet \cite{hou2020multiview} with their original implementations and state-of-the-art methods. On 3 datasets and 4 settings, our implementations outperform the original implementations and achieve competitive results. Although our focus is not on improving these classic architectures, the results indicate that they can still serve as strong baselines.

\begin{table}[]
\centering
\small
\resizebox{\linewidth}{!}{
\begin{tabular}{l|l|c|c}
\toprule
\multirow{2}{*}{MVCNN training} & \multirow{2}{*}{view selection} & \multicolumn{2}{c}{ModelNet40} \\ \cline{3-4}
                                &                                 & 12 views       & 20 views      \\ \hline
from scratch                    & N/A                             & 94.5           & 96.5           \\ \hline
fixed                           & random selection                  & 71.5         & 48.1          \\
fixed                           & dataset-lvl oracle                  & 85.2       & 69.9          \\
fixed                           & instance-lvl oracle                  & 96.5      & 98.1            \\
fixed                           & MVSelect                        & 88.2          & 79.6          \\
joint training                  & MVSelect                        & 94.3           & 94.4         \\ \bottomrule
\end{tabular}
}
\caption{
Evaluation of MVSelect on multiview classification. All view selection results use a total of $T=2$ views. 
}
\label{tab:mvcnn}
\end{table}

\textbf{Experimental setups. }
To demonstrate the full potential of MVSelect, we experiment under two settings. 

First, we fix the task network and only train the view selection module. Our goal is to show that the view selection results outperform random selection. We also compare the performance of two oracles: for a certain initial view, 1) choosing the overall best-performing camera for all instances in the dataset (\textit{dataset-level oracle}) and 2) choosing specifically for each instance (\textit{instance-level oracle}). The former reflects the upper bound of performance achievable by human-designed heuristics, and the latter represents the theoretical performance ceiling when keeping the task network fixed. 


Second, we jointly train the proposed view selection module with the task networks to fully exploit MVSelect. For this experiment, our goal is to achieve the highest possible performance using $T$ views. 


If not specified, we use a total of $T=2$ views for multiview classification and $T=3$ views for detection.

\textbf{Results and analysis.}
In Table~\ref{tab:mvcnn} and Table~\ref{tab:mvdet}, we report the view selection results, which are averaged using all $N$ views as the initial view. 

\begin{figure}
\centering
\includegraphics[width=\linewidth]{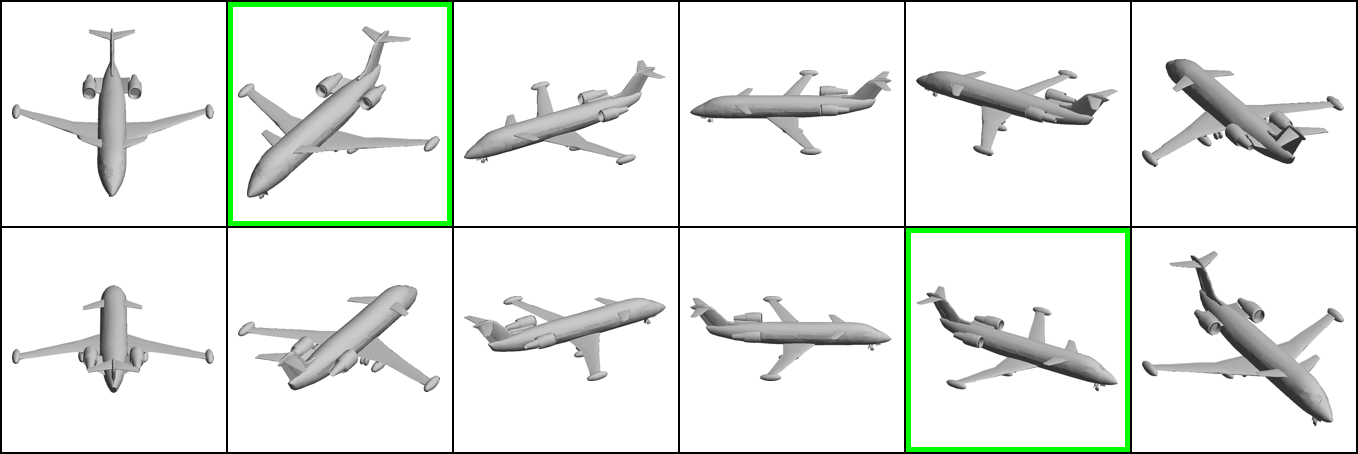}
\caption{Example of selected views on ModelNet40 \cite{wu20153d}.
}
\label{fig:imgs_grid_modelnet}
\end{figure}

\begin{figure}
\centering
    \begin{subfigure}[b]{0.48\linewidth}
    \centering
        \includegraphics[width=\textwidth]{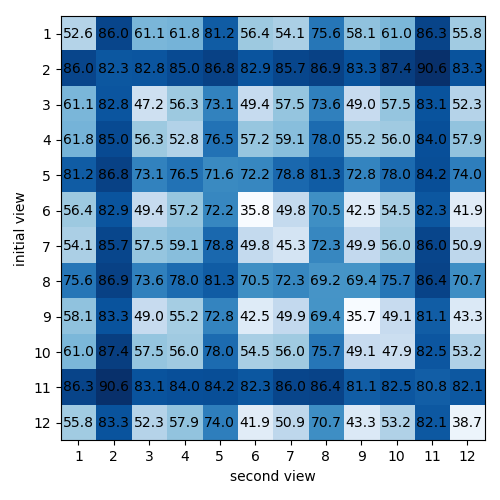}
        \caption{Classification accuracy}
    \end{subfigure}
    \hfill
    \begin{subfigure}[b]{0.48\linewidth}
    \centering
        \includegraphics[width=\textwidth]{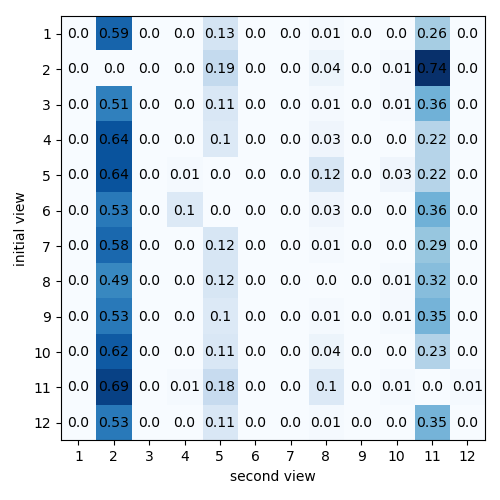}
        \caption{MVSelect policy}
    \end{subfigure}
\caption{Multiview classification with $T=2$ views on the 12-view setup \cite{su2015multi} of ModelNet40 \cite{wu20153d}. The task network is fixed once trained. 
\textbf{Left}: test set accuracy of using two views. \textbf{Right}: MVSelect policy for the test set. 
}
\label{fig:prec_prob}
\end{figure}

For \textbf{multiview classification}, as shown in Table~\ref{tab:mvcnn}, randomly selecting two views cannot achieve competitive results. Specifically, We found that when more views are provided, MVCNN \cite{su2015multi} tends to only activate on the high-confidence views, resulting in worse performance on the 20-view setup than the 12-view setup. 

The proposed view selection module, on the other hand, can choose the supplementary view very effectively. On ModelNet40 dataset \cite{wu20153d}, MVSelect with fixed MVCNN outperforms the random selection baseline by 16.7\% and 31.5\% on the 12-view and 20-view setups, respectively. Compared to dataset-level oracles (same policy for all instances with the same initial camera), MVSelect also turns out to be advantageous by 3.0\% and 9.7\% for the two settings. This verifies that MVSelect can take the target object into consideration (see Fig.~\ref{fig:demo}) and select different cameras for different instances under the same initial camera. 

When we allow joint training for both MVSelect and the task network, we witness large improvements compared to keeping MVCNN fixed. In fact, on two settings, the results are only 0.2\% and 2.1\% behind compared to the full $N$-view system. Overall, we believe that MVSelect and its joint training capabilities enable us to consider only 2 of the 12 or 20 views without major performance drawbacks. 

We demonstrate  an example of the selected views in Fig.~\ref{fig:imgs_grid_modelnet}, and the MVSelect policy in Fig.~\ref{fig:prec_prob}.

For \textbf{multiview detection}, we report view selection results in Table~\ref{tab:mvdet}. 
Since the target scenarios are not fully captured by any individual camera, randomly selecting $T=3$ cameras does not yield satisfactory results. In addition, we observe that the instance-level oracle remains relatively low compared to that of multiview classification tasks (see Table~\ref{tab:mvcnn}). This is likely due to the target scenarios not being fully captured by any single view, and the multiview detection network needs multiple views to collaborate with each other for optimal results. 

In this scenario, we find that MVSelect with a fixed task network can outperform random selection, by 5.1\% MODA on Wildtrack \cite{chavdarova2018wildtrack} and 2.5\% MODA on MultiviewX \cite{hou2020multiview}. Although the raw improvements are not as substantial as those in multiview classification, they are \textbf{statistically highly significant} (p-value $< 0.001$). 
Compared to the dataset-level oracles, we find the MVSelect policy lose its edge. In fact, we find that for multiview detection, MVSelect tends to select the same camera for a given initial view, since it is \textit{not} aware of the situation outside of the FoV coverage. As a result, it cannot choose cameras based on \textit{uncovered} areas in different frames. Without this instance-aware advantage, the fixed camera policy learned during training (MVSelect) cannot outperform the dataset-level oracle, whose policy is computed on the test set.

\begin{table}[]
\centering
\small
\resizebox{\linewidth}{!}{
\begin{tabular}{l|l|c|c}
\toprule
MVDet training & view selection  & Wildtrack       & MultiviewX     \\ \hline
from scratch                    & N/A                             & 90.0           & 93.0          \\ \hline
fixed                           & random selection                  & 74.9         &  76.2           \\
fixed                           & dataset-lvl oracle                  & 82.5       & 80.2           \\
fixed                           & instance-lvl oracle                  & 87.4      & 82.3            \\
fixed                           & MVSelect                        & 80.0           & 78.7          \\
joint training                  & MVSelect                        & 88.6           & 88.1          \\ \bottomrule
\end{tabular}
}
\caption{
Evaluation of MVSelect on multiview detection. We report MODA since it accounts for both precision and recall, plus MODP is not affected too much. All view selection results use a total of $T=3$ views. 
}
\label{tab:mvdet}
\end{table}


\begin{figure}
\centering
\includegraphics[width=\linewidth]{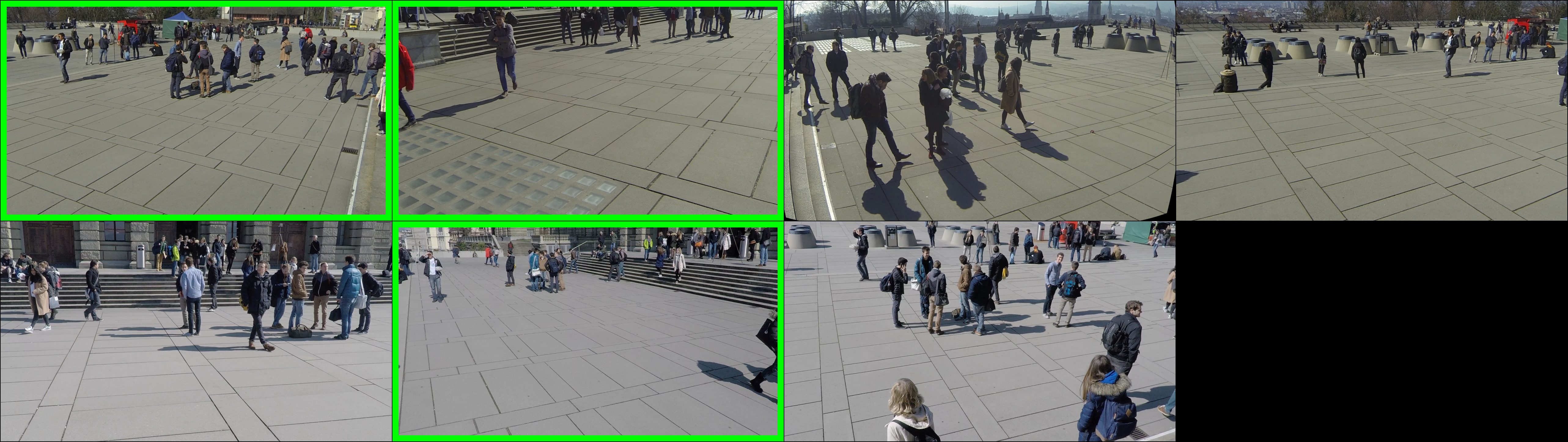}
\caption{Example of selected views on Wildtrack \cite{chavdarova2018wildtrack}.
}
\label{fig:imgs_grid_wildtrack}
\end{figure}

On the positive side, joint training with MVDet once again leads to substantial performance improvements over keeping the task network fixed. In fact, the results even exceed the instance-level oracle for fixed task networks. This phenomenon demonstrates the high level of camera collaboration in normal $N$-camera training protocols, and show that joint training improves task network generalization. Using $T=3$ views, the joint training approach provides competitive results to using all $N$ views, and exceeds the reported performance in the original MVDet paper \cite{hou2020multiview}. 

We present an example of the selected views in the Wildtrack dataset in Fig.~\ref{fig:imgs_grid_wildtrack}.

\begin{figure*}
  \begin{minipage}[t]{0.4\textwidth}
\centering
    \begin{subfigure}[t]{0.48\linewidth}
    \centering
        \includegraphics[width=\textwidth]{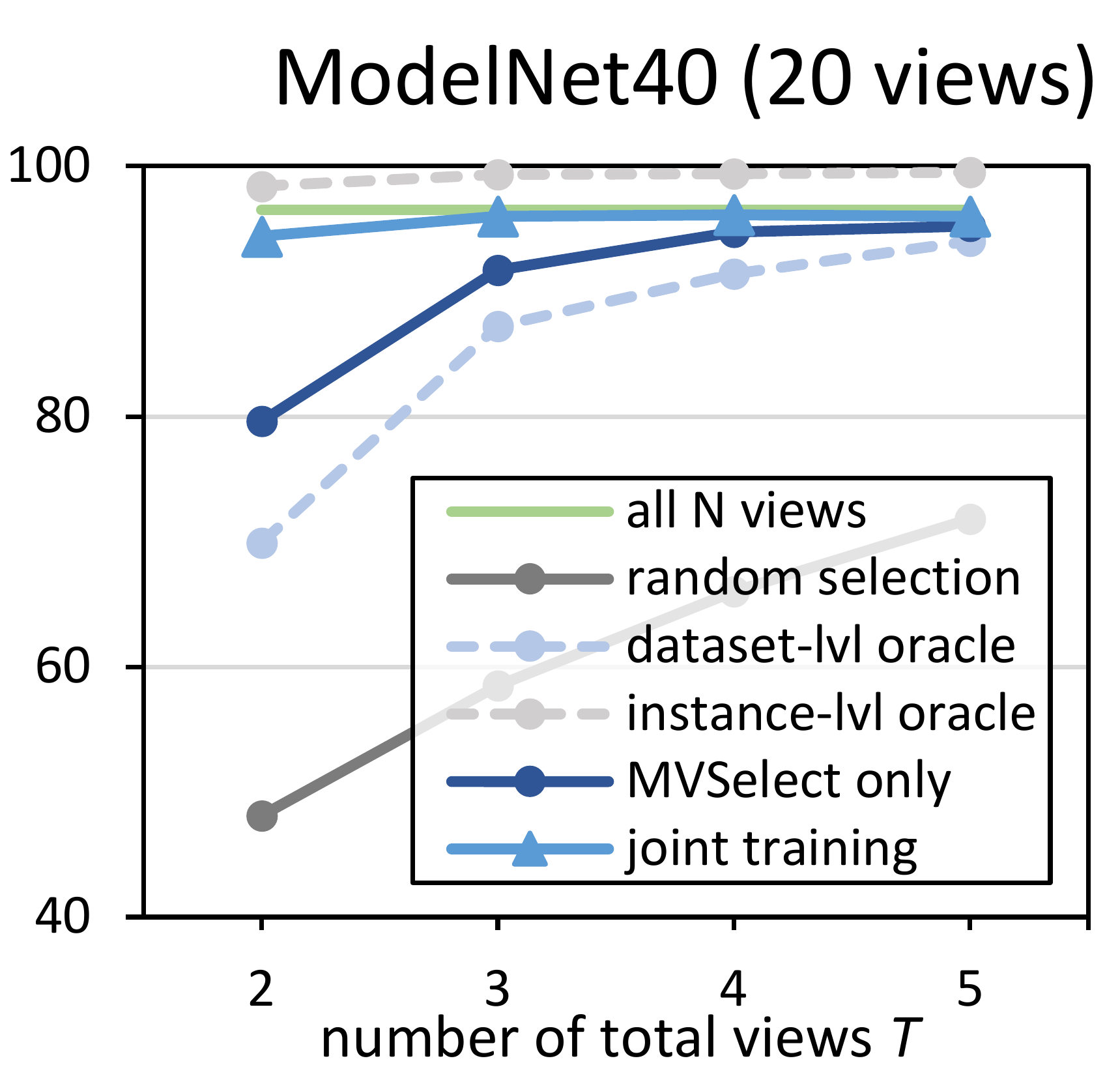}
    \end{subfigure}
    \hfill
    \begin{subfigure}[t]{0.48\linewidth}
    \centering
        \includegraphics[width=\textwidth]{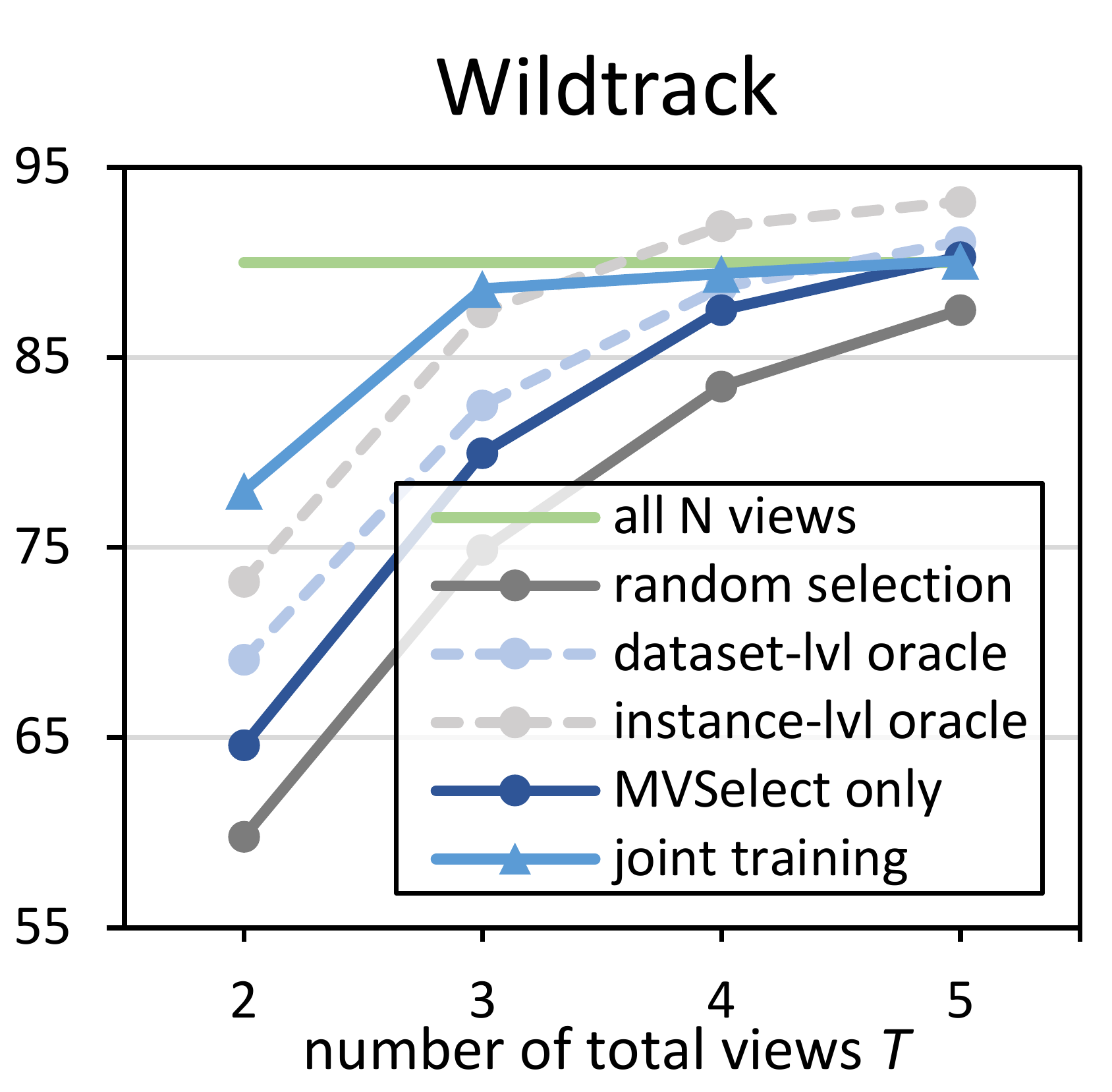}
    \end{subfigure}
\captionof{figure}{System performance under different number of total views $T$. We report classification accuracy and MODA for the two tasks, respectively (same for Fig.~\ref{fig:variants} and Table.~\ref{tab:ablation}). Circle and triangle markers indicate whether the task network is fixed or not, respectively. Dotted lines represent oracle performance (not achievable). 
}
\label{fig:T_curves}
  \end{minipage}
  \hfill
  \begin{minipage}[t]{0.234\textwidth}
    \centering
    \includegraphics[width=\linewidth]{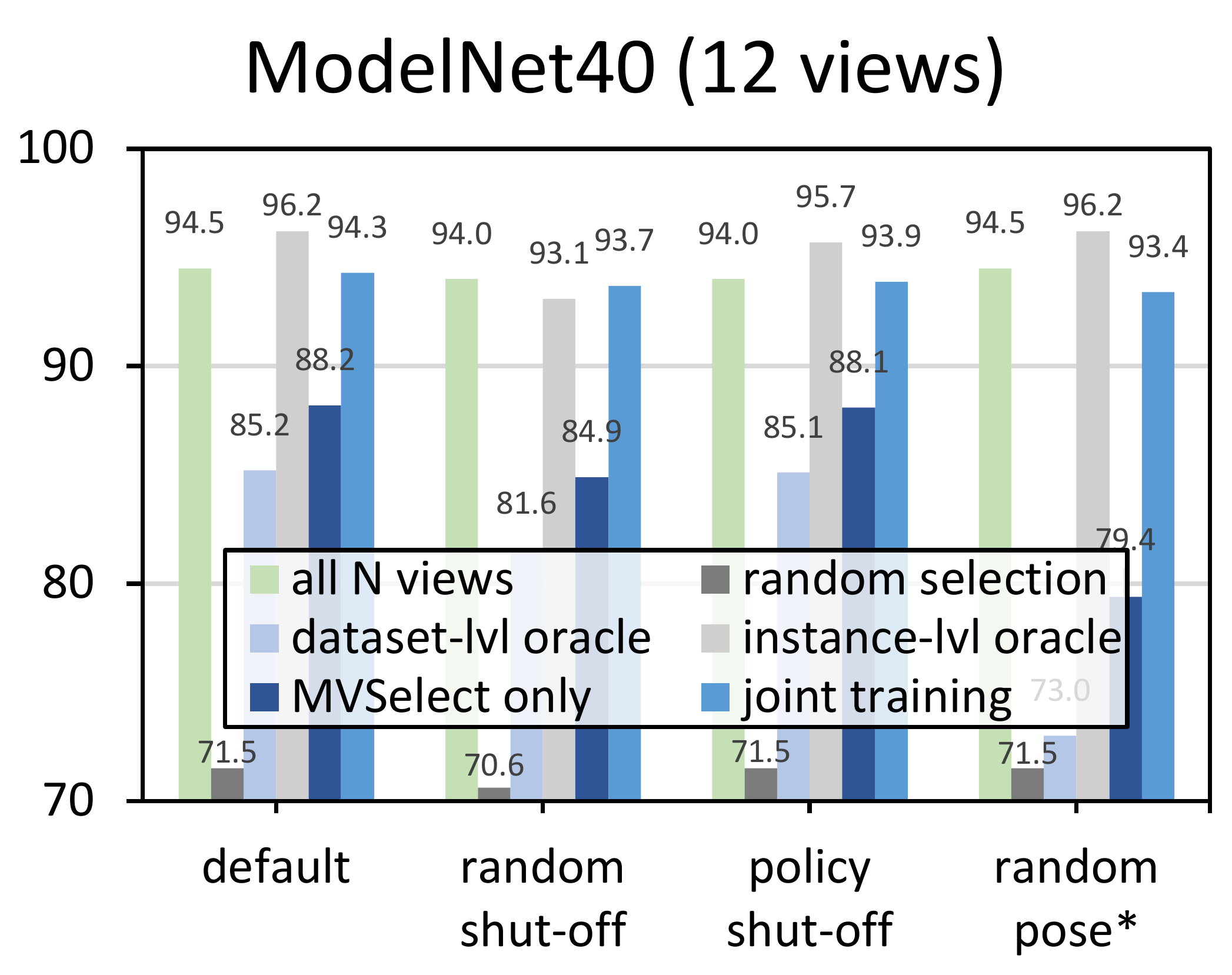}
    \captionof{figure}{System evaluation under different settings. 
    Except for the random pose* setting (where MVSelect variants are re-trained), all models are trained on the default setting.
    }
    \label{fig:variants}
  \end{minipage}
  \hfill
  \begin{minipage}[t]{0.32\textwidth}
\vspace{-32mm}
\centering
\small
\resizebox{\linewidth}{!}{
\setlength{\tabcolsep}{4pt}
\begin{tabular}{l|c|ccc|c}
\toprule
\multirow{2}{*}{}           & \multirow{2}{*}{views} & \multicolumn{3}{c|}{FLOPs}                                        & throughput \\ \cline{3-5}
                            &                        & \multicolumn{1}{c|}{$f\left(\cdot\right)$}      & \multicolumn{1}{c|}{$g\left(\cdot\right)$}     & $d\left(\cdot\right)$      &   (instance/s)                          \\ \hline
\multirow{4}{*}{ModelNet40} & 20                     & \multicolumn{1}{c|}{36.5G}  & \multicolumn{1}{c|}{20.5k} & N/A    & 119.6                       \\ 
                            & 2                      & \multicolumn{1}{c|}{3.6G}   & \multicolumn{1}{c|}{20.5k} & 1.1M   & 361.8                       \\ \cline{2-6} 
                            & 12                     & \multicolumn{1}{c|}{21.9G}  & \multicolumn{1}{c|}{20.5k} & N/A    & 196.2                       \\ 
                            & 2                      & \multicolumn{1}{c|}{3.6G}   & \multicolumn{1}{c|}{20.5k} & 530.4k & 507.9                       \\ \hline
\multirow{2}{*}{Wildtrack}  & 7                      & \multicolumn{1}{c|}{1.2T}   & \multicolumn{1}{c|}{19.1G} & N/A    & 8.4                         \\ 
                            & 3                      & \multicolumn{1}{c|}{511.6G} & \multicolumn{1}{c|}{19.1G} & 3.2G   & 16.2                        \\ \hline
\multirow{2}{*}{MultiviewX} & 6                      & \multicolumn{1}{c|}{1.0T}   & \multicolumn{1}{c|}{17.7G} & N/A    & 9.8                         \\ 
                            & 3                      & \multicolumn{1}{c|}{511.6G} & \multicolumn{1}{c|}{17.7G} & 2.9G   & 16.9                        \\ \bottomrule
\end{tabular}
}
\captionof{table}{Computational cost and speed.}
\label{tab:efficiency}
\vspace{4mm}
\centering
\small
\resizebox{\linewidth}{!}{
\begin{tabular}{l|c|c}
\toprule
\multirow{2}{*}{}  & ModelNet40 & \multirow{2}{*}{Wildtrack} \\ \cline{2-2}
                   & 12 views   &                            \\ \hline
MVSelect only      & 88.2                  & 80.0      \\
w/o camera branch  & 88.2                  & 79.1      \\
w/o feature branch & 85.0                  & 79.9      \\ \bottomrule
\end{tabular}
}
\captionof{table}{Ablation study. }
\label{tab:ablation}
  \end{minipage}
\end{figure*}

\subsection{Efficiency Analysis}
In Table~\ref{tab:efficiency}, following previous works in efficient inference \cite{li2017pruning,howard2017mobilenets}, we detail the computational cost in FLOPs for task networks and MVSelect. Specifically, we find feature extraction $f\left(\cdot\right)$ to take up the majority of the computation, while everything else is lighter by at least an order of magnitude. Overall, we verify that using $T=2$ or 3 out of $N$ views can reduce the computational cost to roughly $\nicefrac{T}{N}$. 

In terms of inference speed, we find reduction in FLOPs results in monotonically increasing throughputs, ranging from $1.72\times$ to $3.03\times$. Due to factors such as implementation, parallelization, and hardware limitations, actual speedups cannot actually reach the level of computational cost reduction, as suggested by previous study \cite{molchanov2016pruning}. 

\subsection{Variant Study}

\textbf{Influence of total view count.}
The performance curves in Fig.~\ref{fig:T_curves} demonstrate that the joint training variant achieves competitive performance with as few as $T=2$ views for multiview classification and $T=3$ views for multiview detection, beyond which point performance plateaus. By contrast, learning MVSelect for fixed task networks shows a less steep curve. Notably, both datasets exhibit an increase in performance up to a total of $T=5$ views, at which point the MVSelect policy performs comparably to the full $N$-view system.
\textbf{Camera layout optimization.}
Determining the optimal camera locations is crucial for setting up an effective multiview system. In Fig.\ref{fig:prec_prob}, we observe that not all cameras are equally useful according to the MVSelect policy. To address this, we allocate a validation partition of the data to identify the more useful camera views and then disable half of the $N=12$ cameras that are not frequently utilized. In testing, we find that this new camera layout (policy shut-off'' in Fig.~\ref{fig:variants}) does not significantly impact the task network or the MVSelect module, and outperforms randomly disabling 6 of the 12 cameras (random shut-off'' in Fig.\ref{fig:variants}).
Although it is necessary to set up all $N$ cameras for the analysis, optimizing the multiview camera layout can be a crucial step towards achieving optimal performance, and merits further investigation.


\textbf{Random object pose.}
In real-world applications of multiview systems, such as those found in iPhones and Teslas, the cameras may remain fixed in their relative positions while the entire system is in motion. To simulate this scenario in our experiments, we introduce the random object pose setting (Fig.~\ref{fig:variants}) and re-train MVSelect. While there is no exact object pose as supervision, the reinforcement learning approach is able to roughly infer the relative poses between the object and the multiview system, resulting in improved performance compared to random selection and dataset-level oracle (which is arguably inappropriate for this setup). In the future, we plan to investigate the use of relative poses between multiview systems and the environment as supervision for moving system setups.

\textbf{Ablation study.}
Regarding the MVSelect architecture design, Table~\ref{tab:ablation} shows that removing the camera branch results in a MODA drop for the Wildtrack dataset, while the classification accuracy on ModelNet40 is not affected. Removing the feature branch brings the classification accuracy to dataset-level oracle (Table~\ref{tab:mvcnn}), but it does not significantly affect the multiview classification results. 

This aligns with the policy we learned for the two tasks. For multiview detection, since the system has no clue about areas outside camera FoVs, MVSelect tends to guess and choose cameras that provide the overall best results. Therefore, the camera branch plays a more important role, as it encodes prior knowledge based on the initial camera. For multiview classification, however, target objects are fully observable, and MVSelect can make different decisions for each instance. In this case, the feature branch is more important, as it enables per-instance decision making. 



\section{Conclusion}


In conclusion, this paper proposes an efficient approach for multiview systems by leveraging the camera layout and limiting the number of views. Through the use of a camera view selection module, MVSelect, jointly trained with the task network using reinforcement learning, the proposed approach demonstrates competitive performance on multiview classification and detection tasks at fractions of the computational cost. Overall, the proposed efficient approach provides an alternative to reducing image resolution and using lighter networks, and paves way for future multiview camera layout optimization.

{\small
\bibliographystyle{ieee_fullname}
\bibliography{egbib}

\begin{thebibliography}{10}\itemsep=-1pt

\bibitem{aloimonos1988active}
John Aloimonos, Isaac Weiss, and Amit Bandyopadhyay.
\newblock Active vision.
\newblock {\em International journal of computer vision}, 1:333--356, 1988.

\bibitem{anderson2018vision}
Peter Anderson, Qi Wu, Damien Teney, Jake Bruce, Mark Johnson, Niko
  S{\"u}nderhauf, Ian Reid, Stephen Gould, and Anton Van Den~Hengel.
\newblock Vision-and-language navigation: Interpreting visually-grounded
  navigation instructions in real environments.
\newblock In {\em Proceedings of the IEEE conference on computer vision and
  pattern recognition}, pages 3674--3683, 2018.

\bibitem{iphone14pro}
Apple.
\newblock iphone 14 pro and iphone 14 pro max.
\newblock \url{https://www.apple.com/iphone-14-pro/}.

\bibitem{baque2017deep}
Pierre Baqu{\'e}, Fran{\c{c}}ois Fleuret, and Pascal Fua.
\newblock Deep occlusion reasoning for multi-camera multi-target detection.
\newblock In {\em Proceedings of the IEEE International Conference on Computer
  Vision}, pages 271--279, 2017.

\bibitem{chavdarova2018wildtrack}
Tatjana Chavdarova, Pierre Baqu{\'e}, St{\'e}phane Bouquet, Andrii Maksai, Cijo
  Jose, Timur Bagautdinov, Louis Lettry, Pascal Fua, Luc Van~Gool, and
  Fran{\c{c}}ois Fleuret.
\newblock Wildtrack: A multi-camera hd dataset for dense unscripted pedestrian
  detection.
\newblock In {\em Proceedings of the IEEE Conference on Computer Vision and
  Pattern Recognition}, pages 5030--5039, 2018.

\bibitem{chavdarova2017deep}
Tatjana Chavdarova et~al.
\newblock Deep multi-camera people detection.
\newblock In {\em 2017 16th IEEE International Conference on Machine Learning
  and Applications (ICMLA)}, pages 848--853. IEEE, 2017.

\bibitem{chen2011active}
Shengyong Chen, Youfu Li, and Ngai~Ming Kwok.
\newblock Active vision in robotic systems: A survey of recent developments.
\newblock {\em The International Journal of Robotics Research},
  30(11):1343--1377, 2011.

\bibitem{dosovitskiy2020vit}
Alexey Dosovitskiy, Lucas Beyer, Alexander Kolesnikov, Dirk Weissenborn,
  Xiaohua Zhai, Thomas Unterthiner, Mostafa Dehghani, Matthias Minderer, Georg
  Heigold, Sylvain Gelly, Jakob Uszkoreit, and Neil Houlsby.
\newblock An image is worth 16x16 words: Transformers for image recognition at
  scale.
\newblock {\em ICLR}, 2021.

\bibitem{smart_city_camera}
ABI Research: The Tech~Intelligence Experts.
\newblock More than 155,000 smart ai-based cameras will transform traffic
  management by 2025.
\newblock
  \url{https://www.abiresearch.com/press/more-155000-smart-ai-based-cameras-will-transform-traffic-management-2025/}.

\bibitem{feng2018gvcnn}
Yifan Feng, Zizhao Zhang, Xibin Zhao, Rongrong Ji, and Yue Gao.
\newblock Gvcnn: Group-view convolutional neural networks for 3d shape
  recognition.
\newblock In {\em Proceedings of the IEEE conference on computer vision and
  pattern recognition}, pages 264--272, 2018.

\bibitem{findlay2003active}
John~M Findlay and Iain~D Gilchrist.
\newblock {\em Active vision: The psychology of looking and seeing}.
\newblock Number~37. Oxford University Press, 2003.

\bibitem{fleuret2007multicamera}
Francois Fleuret, Jerome Berclaz, Richard Lengagne, and Pascal Fua.
\newblock Multicamera people tracking with a probabilistic occupancy map.
\newblock {\em IEEE transactions on pattern analysis and machine intelligence},
  30(2):267--282, 2007.

\bibitem{gao2022exploiting}
Xin Gao, Yijin Xiong, Guoying Zhang, Hui Deng, and Kangkang Kou.
\newblock Exploiting key points supervision and grouped feature fusion for
  multiview pedestrian detection.
\newblock {\em Pattern Recognition}, 131:108866, 2022.

\bibitem{hamdi2021mvtn}
Abdullah Hamdi, Silvio Giancola, and Bernard Ghanem.
\newblock Mvtn: Multi-view transformation network for 3d shape recognition.
\newblock In {\em Proceedings of the IEEE/CVF International Conference on
  Computer Vision}, pages 1--11, 2021.

\bibitem{he2016deep}
Kaiming He, Xiangyu Zhang, Shaoqing Ren, and Jian Sun.
\newblock Deep residual learning for image recognition.
\newblock In {\em Proceedings of the IEEE conference on computer vision and
  pattern recognition}, pages 770--778, 2016.

\bibitem{hou2021multiview}
Yunzhong Hou and Liang Zheng.
\newblock Multiview detection with shadow transformer (and view-coherent data
  augmentation).
\newblock In {\em Proceedings of the 29th ACM International Conference on
  Multimedia}, pages 1673--1682, 2021.

\bibitem{hou2020multiview}
Yunzhong Hou, Liang Zheng, and Stephen Gould.
\newblock Multiview detection with feature perspective transformation.
\newblock In {\em European Conference on Computer Vision}, pages 1--18.
  Springer, 2020.

\bibitem{howard2017mobilenets}
Andrew~G Howard, Menglong Zhu, Bo Chen, Dmitry Kalenichenko, Weijun Wang,
  Tobias Weyand, Marco Andreetto, and Hartwig Adam.
\newblock Mobilenets: Efficient convolutional neural networks for mobile vision
  applications.
\newblock {\em arXiv preprint arXiv:1704.04861}, 2017.

\bibitem{hwang2022booster}
Jinwoo Hwang, Philipp Benz, and Tae-hoon Kim.
\newblock Booster-shot: Boosting stacked homography transformations for
  multiview pedestrian detection with attention.
\newblock {\em arXiv preprint arXiv:2208.09211}, 2022.

\bibitem{kanezaki2018rotationnet}
Asako Kanezaki, Yasuyuki Matsushita, and Yoshifumi Nishida.
\newblock Rotationnet: Joint object categorization and pose estimation using
  multiviews from unsupervised viewpoints.
\newblock In {\em Proceedings of the IEEE conference on computer vision and
  pattern recognition}, pages 5010--5019, 2018.

\bibitem{kasturi2008framework}
Rangachar Kasturi, Dmitry Goldgof, Padmanabhan Soundararajan, Vasant Manohar,
  John Garofolo, Rachel Bowers, Matthew Boonstra, Valentina Korzhova, and Jing
  Zhang.
\newblock Framework for performance evaluation of face, text, and vehicle
  detection and tracking in video: Data, metrics, and protocol.
\newblock {\em IEEE Transactions on Pattern Analysis and Machine Intelligence},
  31(2):319--336, 2008.

\bibitem{kingma2015adam}
Diederik~P Kingma and Jimmy Ba.
\newblock Adam: A method for stochastic optimization.
\newblock In {\em ICLR (Poster)}, 2015.

\bibitem{kipf2016semi}
Thomas~N Kipf and Max Welling.
\newblock Semi-supervised classification with graph convolutional networks.
\newblock In {\em J. International Conference on Learning Representations (ICLR
  2017)}, 2016.

\bibitem{li2017pruning}
Hao Li, Asim Kadav, Igor Durdanovic, Hanan Samet, and Hans~Peter Graf.
\newblock Pruning filters for efficient convnets.
\newblock In {\em International Conference on Learning Representations}, 2017.

\bibitem{lima20223d}
Jo{\~a}o~Paulo Lima, Rafael Roberto, Lucas Figueiredo, Francisco Sim{\~o}es,
  Diego Thomas, Hideaki Uchiyama, and Veronica Teichrieb.
\newblock 3d pedestrian localization using multiple cameras: a generalizable
  approach.
\newblock {\em Machine Vision and Applications}, 33(4):1--16, 2022.

\bibitem{mnih2014recurrent}
Volodymyr Mnih, Nicolas Heess, Alex Graves, et~al.
\newblock Recurrent models of visual attention.
\newblock {\em Advances in neural information processing systems}, 27, 2014.

\bibitem{mnih2013playing}
Volodymyr Mnih, Koray Kavukcuoglu, David Silver, Alex Graves, Ioannis
  Antonoglou, Daan Wierstra, and Martin Riedmiller.
\newblock Playing atari with deep reinforcement learning.
\newblock {\em arXiv preprint arXiv:1312.5602}, 2013.

\bibitem{molchanov2016pruning}
Pavlo Molchanov, Stephen Tyree, Tero Karras, Timo Aila, and Jan Kautz.
\newblock Pruning convolutional neural networks for resource efficient
  inference.
\newblock {\em arXiv preprint arXiv:1611.06440}, 2016.

\bibitem{argo}
University of Parma and University of Pavia.
\newblock The argo project.
\newblock \url{http://www.argo.ce.unipr.it/ARGO/english/flyer_en.pdf}.

\bibitem{qi2016volumetric}
Charles~R Qi, Hao Su, Matthias Nie{\ss}ner, Angela Dai, Mengyuan Yan, and
  Leonidas~J Guibas.
\newblock Volumetric and multi-view cnns for object classification on 3d data.
\newblock In {\em Proceedings of the IEEE conference on computer vision and
  pattern recognition}, pages 5648--5656, 2016.

\bibitem{qiu20223d}
Rui Qiu, Ming Xu, Yuyao Yan, Jeremy~S Smith, and Xi Yang.
\newblock 3d random occlusion and multi-layer projection for deep multi-camera
  pedestrian localization.
\newblock {\em arXiv preprint arXiv:2207.10895}, 2022.

\bibitem{roig2011conditional}
Gemma Roig, Xavier Boix, Horesh~Ben Shitrit, and Pascal Fua.
\newblock Conditional random fields for multi-camera object detection.
\newblock In {\em 2011 International Conference on Computer Vision}, pages
  563--570. IEEE, 2011.

\bibitem{sabottke2020effect}
Carl~F Sabottke and Bradley~M Spieler.
\newblock The effect of image resolution on deep learning in radiography.
\newblock {\em Radiology. Artificial intelligence}, 2(1), 2020.

\bibitem{schulman2015gradient}
John Schulman, Nicolas Heess, Theophane Weber, and Pieter Abbeel.
\newblock Gradient estimation using stochastic computation graphs.
\newblock {\em Advances in neural information processing systems}, 28, 2015.

\bibitem{schulman2017proximal}
John Schulman, Filip Wolski, Prafulla Dhariwal, Alec Radford, and Oleg Klimov.
\newblock Proximal policy optimization algorithms.
\newblock {\em arXiv preprint arXiv:1707.06347}, 2017.

\bibitem{shermeyer2019effects}
Jacob Shermeyer and Adam Van~Etten.
\newblock The effects of super-resolution on object detection performance in
  satellite imagery.
\newblock In {\em Proceedings of the IEEE/CVF Conference on Computer Vision and
  Pattern Recognition Workshops}, pages 0--0, 2019.

\bibitem{song2021stacked}
Liangchen Song, Jialian Wu, Ming Yang, Qian Zhang, Yuan Li, and Junsong Yuan.
\newblock Stacked homography transformations for multi-view pedestrian
  detection.
\newblock In {\em Proceedings of the IEEE/CVF International Conference on
  Computer Vision}, pages 6049--6057, 2021.

\bibitem{su2015multi}
Hang Su, Subhransu Maji, Evangelos Kalogerakis, and Erik Learned-Miller.
\newblock Multi-view convolutional neural networks for 3d shape recognition.
\newblock In {\em Proceedings of the IEEE international conference on computer
  vision}, pages 945--953, 2015.

\bibitem{sutton2018reinforcement}
Richard~S Sutton and Andrew~G Barto.
\newblock {\em Reinforcement learning: An introduction}.
\newblock MIT press, 2018.

\bibitem{unity}
Unity Technologies.
\newblock Unity.
\newblock \url{https://unity.com/}.

\bibitem{tesla_model3_manual}
Tesla.
\newblock Model 3 owner's manual.
\newblock
  \url{https://www.tesla.com/ownersmanual/model3/en_us/GUID-EDA77281-42DC-4618-98A9-CC62378E0EC2.html}.

\bibitem{vora2021bringing}
Jeet Vora, Swetanjal Dutta, Kanishk Jain, Shyamgopal Karthik, and Vineet
  Gandhi.
\newblock Bringing generalization to deep multi-view detection.
\newblock {\em arXiv preprint arXiv:2109.12227}, 2021.

\bibitem{wang2019dominant}
Chu Wang, Marcello Pelillo, and Kaleem Siddiqi.
\newblock Dominant set clustering and pooling for multi-view 3d object
  recognition.
\newblock {\em arXiv preprint arXiv:1906.01592}, 2019.

\bibitem{wei2020view}
Xin Wei, Ruixuan Yu, and Jian Sun.
\newblock View-gcn: View-based graph convolutional network for 3d shape
  analysis.
\newblock In {\em Proceedings of the IEEE/CVF Conference on Computer Vision and
  Pattern Recognition}, pages 1850--1859, 2020.

\bibitem{williams1992simple}
Ronald~J Williams.
\newblock Simple statistical gradient-following algorithms for connectionist
  reinforcement learning.
\newblock {\em Reinforcement learning}, pages 5--32, 1992.

\bibitem{wu20153d}
Zhirong Wu, Shuran Song, Aditya Khosla, Fisher Yu, Linguang Zhang, Xiaoou Tang,
  and Jianxiong Xiao.
\newblock 3d shapenets: A deep representation for volumetric shapes.
\newblock In {\em Proceedings of the IEEE conference on computer vision and
  pattern recognition}, pages 1912--1920, 2015.

\bibitem{xu2016multi}
Yuanlu Xu, Xiaobai Liu, Yang Liu, and Song-Chun Zhu.
\newblock Multi-view people tracking via hierarchical trajectory composition.
\newblock In {\em Proceedings of the IEEE Conference on Computer Vision and
  Pattern Recognition}, pages 4256--4265, 2016.

\bibitem{yang2019learning}
Ze Yang and Liwei Wang.
\newblock Learning relationships for multi-view 3d object recognition.
\newblock In {\em Proceedings of the IEEE/CVF International Conference on
  Computer Vision}, pages 7505--7514, 2019.

\bibitem{yu2018multi}
Tan Yu, Jingjing Meng, and Junsong Yuan.
\newblock Multi-view harmonized bilinear network for 3d object recognition.
\newblock In {\em Proceedings of the IEEE conference on computer vision and
  pattern recognition}, pages 186--194, 2018.

\bibitem{zhu2017target}
Yuke Zhu, Roozbeh Mottaghi, Eric Kolve, Joseph~J Lim, Abhinav Gupta, Li
  Fei-Fei, and Ali Farhadi.
\newblock Target-driven visual navigation in indoor scenes using deep
  reinforcement learning.
\newblock In {\em 2017 IEEE international conference on robotics and automation
  (ICRA)}, pages 3357--3364. IEEE, 2017.

\end{thebibliography}
}

\end{document}